%% file: codepmp.tex
\pdfoutput=1

\documentclass[11pt]{article}

\usepackage{EMNLP2023}

\usepackage{times}
\usepackage{latexsym}

\usepackage[T1]{fontenc}
\usepackage[utf8]{inputenc}
\usepackage{microtype}
\usepackage{inconsolata}

\usepackage{graphicx}
\usepackage{booktabs}
\usepackage{hyperref}
\usepackage{amsmath}
\usepackage{amssymb}
\usepackage{mathtools}
\usepackage{amsthm}
\usepackage[capitalize,noabbrev]{cleveref}
\usepackage{enumitem}
\input{math_commands.tex}
\usepackage{multirow}
\usepackage{multicol}
\usepackage{pifont}
\usepackage{xcolor}
\usepackage{soul}
\usepackage{url}
\usepackage{wrapfig}
\usepackage{makecell}
\usepackage{adjustbox}
\usepackage{siunitx}
\usepackage{lipsum}
\usepackage{subcaption}
\usepackage{eqparbox}
\usepackage{longtable}
\usepackage{algorithm}
\usepackage{algorithmic}
\usepackage{marvosym}

\setlist{topsep=2pt, partopsep=0pt, parsep=0pt, itemsep=2pt}

\theoremstyle{plain}

\theoremstyle{definition}

\theoremstyle{remark}

\title{CodePMP: Scalable Preference Model Pretraining for Large Language Model Reasoning}

\author{
   Huimu Yu\textsuperscript{\rm 1,2}, Xing Wu\thanks{Equal contribution. This work is jointly completed by Huimu Yu under the guidance of Xing Wu.}\textsuperscript{\rm 1,2,3 \Letter}, Haotian Xu\textsuperscript{\rm 3}, Debing Zhang\textsuperscript{\rm 3}, 
   \textbf{Songlin Hu}\textsuperscript{\rm 1,2 \Letter}
  \\
  \textsuperscript{\rm 1}Institute of Information Engineering, Chinese Academy of Sciences\\
  \textsuperscript{\rm 2}School of Cyber Security, University of Chinese Academy of Sciences\\
  \textsuperscript{\rm 3}Xiaohongshu Inc\\
  \{yuhunmu,wuxing,husonglin\}@iie.ac.cn, \{xuhaotian,dengyang\}@xiaohongshu.com
}

\begin{document}
\maketitle

\begin{abstract}
Large language models (LLMs) have made significant progress in natural language understanding and generation, driven by scalable pretraining and advanced finetuning. However, enhancing reasoning abilities in LLMs, particularly via reinforcement learning from human feedback (RLHF), remains challenging due to the scarcity of high-quality preference data, which is labor-intensive to annotate and crucial for reward model (RM) finetuning. To alleviate this issue, we introduce CodePMP, a scalable preference model pretraining (PMP) pipeline that utilizes a large corpus of synthesized code-preference pairs from publicly available high-quality source code. CodePMP improves RM finetuning efficiency by pretraining preference models on large-scale synthesized code-preference pairs. We evaluate CodePMP on mathematical reasoning tasks (GSM8K, MATH) and logical reasoning tasks (ReClor, LogiQA2.0), consistently showing significant improvements in reasoning performance of LLMs and highlighting the importance of scalable preference model pretraining for efficient reward modeling.
\end{abstract}

\section{Introduction}
Large language models (LLMs) have achieved remarkable progress in natural language understanding and generation, driven by advancements in scalable pretraining and finetuning techniques, including supervised finetuning (SFT)~\citep{wang2022self,wang2023making} and Reinforcement Learning from Human Feedback (RLHF)~\citep{bai2022training,lightman2023let,bai2022constitutional,gulcehre2023reinforced,schulman2017proximal,rafailov2024direct}. Despite these advances, enhancing LLMs' reasoning capabilities, particularly for complex logical and mathematical tasks, remains a significant challenge~\citep{wang2023math,zhang2024unveiling}. While RLHF has proven effective for improving model performance, its efficacy is constrained by the availability of high-quality preference data, which is expensive and labor-intensive to collect~\citep{cobbe2021training,zheng2024judging}. This limitation impedes the scalability of reward model (RM) finetuning, which is instrumental in guiding LLMs toward optimal outputs.

\begin{figure}[t]
\vspace{-4pt}
\centering
\includegraphics[width=\columnwidth]{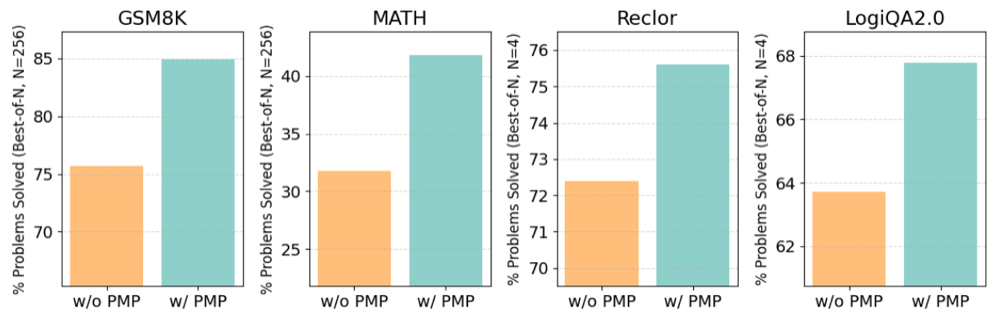}
\caption{Compared to directly finetuning reward models, CodePMP significantly improves the sample efficiency and capability of reward models, which in turn boosts the generator's(MetaMath-Mistral-7B) reasoning performance (Best-of-N accuracy) across both mathematical reasoning tasks (GSM8K and MATH) and logical reasoning tasks (ReClor and LogiQA2.0).}
\label{fig:codepmp_highlight}
\end{figure}

To alleviate this issue, prior works like Anthropic's Preference Model Pretraining (PMP)~\citep{askell2021general} have proposed improving reward modeling data efficiency by pretraining preference models on large-scale preference data from public sources like Reddit and Wikipedia, followed by an efficient finetuning on limited high-quality human-annotated data. However, this approach is less effective for reasoning tasks due to the scarcity of reasoning preference pairs available online. Compared to other tasks, manually annotating preference data for reasoning is inherently more challenging to scale~\citep{zhang2024unveiling,zhou2023solving}, highlighting the urgent need for a scalable PMP approach for reasoning tasks.

\begin{figure*}[t]
\vspace{-8pt}
\begin{center}
\centerline{\includegraphics[width=2\columnwidth]{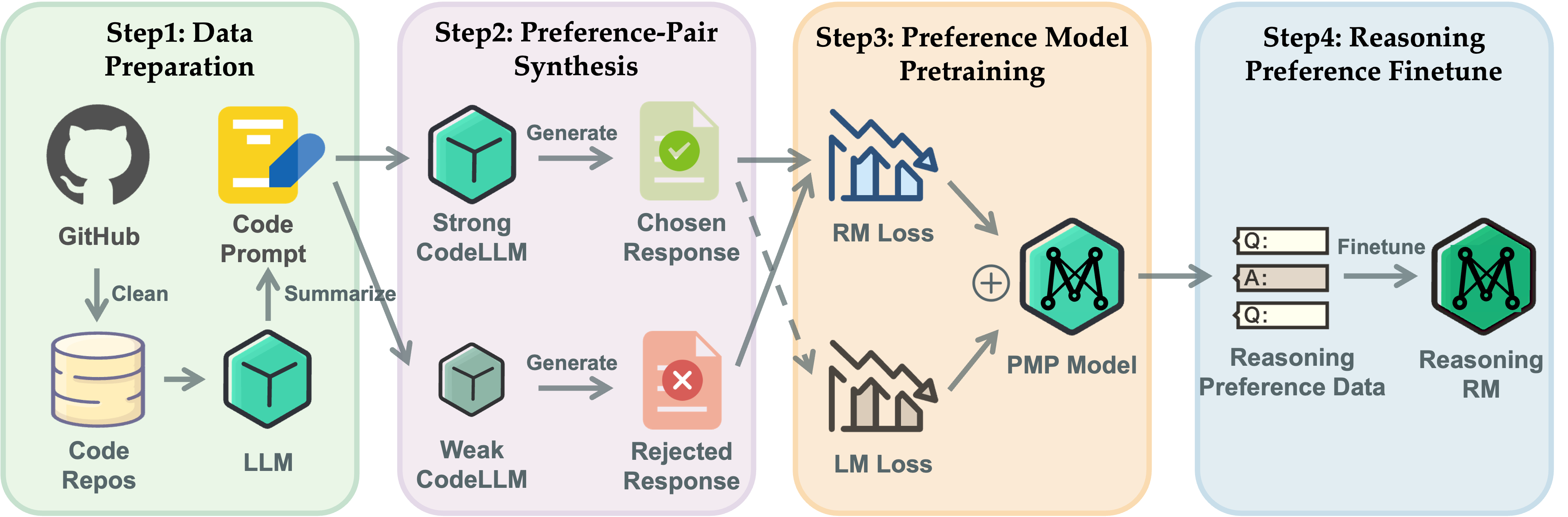}}
\caption{\textbf{Overview of CodePMP.} First, raw code collected from GitHub is cleaned and summarized into code prompts (descriptions). Then, a weak CodeLLM generates \textit{rejected} responses while a stronger CodeLLM produces \textit{chosen} responses. Finally, these millions of $\langle \textit{chosen}, \textit{rejected} \rangle$ pairs form the preference model pretraining dataset, enhancing both sample efficiency and performance for downstream reasoning tasks.}
\label{fig:codepmp_arch_new}
\end{center}
\vspace{-20pt}
\end{figure*}

In this paper, we propose \textbf{CodePMP}, a scalable preference model pretraining pipeline that enhances LLM reasoning abilities using synthesized preference pairs derived from high-quality, publicly available source code. Code, with its inherently logical and structured nature, provides rich data suitable for reasoning tasks. Recent works~\citep{zhang2024unveiling,aryabumi2024code} also show a strong correlation between code training and reasoning improvements in LLMs. By leveraging the huge amount and diverse coverage of source code available on platforms like GitHub, CodePMP offers a scalable solution for pretraining preference models, thereby improving RM finetuning efficiency and enhancing LLMs' reasoning performance.

Specifically, CodePMP generates preference pairs by synthesizing \textit{chosen} and \textit{rejected} code responses for a given code-related prompt or description using CodeLLMs. A strong CodeLLM produces higher-quality (\textit{chosen}) responses, while a weaker model generates suboptimal or even low-quality (\textit{rejected}) responses. These $\langle \textit{chosen}, \textit{rejected} \rangle$ pairs, accumulated in the millions, form a large-scale synthesized preference dataset. This dataset is then used to pretrain the preference model with pairwise ranking objectives~\citep{cobbe2021training,charniak2005coarse}, providing a good initialization for further finetuning the reward models.

We evaluate CodePMP on widely studied reasoning tasks, including mathematical reasoning tasks such as GSM8K~\citep{cobbe2021training} and MATH~\citep{hendrycks2021measuring}, as well as logical reasoning tasks like ReClor~\citep{yu2020reclor} and LogiQA2.0~\citep{liu2023logiqa}. Our experiments show that CodePMP significantly improves RM finetuning accuracy and Best-of-N performance in reasoning tasks, outperforming direct RM finetuning, as highlighted in Figure~\ref{fig:codepmp_highlight}. Moreover, additional results reveal that RMs initialized with CodePMP exhibit greater robustness across different tasks. These results indicate that code-derived preference data provides a scalable, cost-effective solution for enhancing LLM reasoning capabilities while reducing reliance on extensive preference annotation, achieving more effective reward modeling for reasoning tasks.

In summary, our main contributions are:
\begin{enumerate}[leftmargin=*]
    \item We introduce CodePMP, a scalable method that uses code-derived preference pairs to pretrain preference models, improving sample efficiency and robustness for downstream RM finetuning.
    \item We validate that CodePMP significantly improves performance on reasoning tasks, demonstrating that a scalable PMP process positively impacts LLM reasoning abilities.
    \item We provide a detailed analysis of key design elements in CodePMP, offering valuable insights for future research in related areas.
\end{enumerate}

\section{Preliminaries}

\paragraph{Language Modeling} Language modeling represents a fundamental task in natural language processing aimed at modeling sequential language data. This is typically implemented through Causal Language Models (Causal LM), which maximize the likelihood of predicting the next token $w_t$ given preceding tokens $w_1, w_2, \dots, w_{t-1}$. The training process minimizes the negative log-likelihood:
\begin{equation}
\mathcal{L}_{\text{LM}} = - \sum_{t=1}^{T} \log P(w_t | w_1, w_2, \dots, w_{t-1})
\end{equation}
This loss function $\mathcal{L}_{\text{LM}}$ encourages the model to capture underlying patterns in the data. Transformer architectures~\citep{vaswani2017attention} are the standard for Causal LM due to their ability to handle long-range dependencies effectively.

\paragraph{Reward Modeling} Reward modeling (RM) is integral to reinforcement learning from human feedback (RLHF), providing scalar reward signals that guide learning based on output quality. The reward model $R_\theta$ predicts the quality of an output $y$ given a context $x$ as $s = R_\theta(x, y)$. In preference modeling, RMs predict relative quality by comparing output pairs. A standard approach employs the Pairwise Ranking Loss, which assigns higher scores to preferred (chosen) outputs:
\begin{equation}
\mathcal{L}_{\text{RM}} = -\log\left(\sigma(s_{\text{chosen}} - s_{\text{rejected}})\right)
\end{equation}
\noindent, where $s_{\text{chosen}} = R_\theta(x, y_{\text{chosen}})$ and $s_{\text{rejected}} = R_\theta(x, y_{\text{rejected}})$, and $\sigma(\cdot)$ is the sigmoid function.

\paragraph{Best-of-N Sampling} Best-of-N (BoN) sampling enhances LLM reasoning~\citep{cobbe2021training, lightman2023let} by generating $N$ candidate solutions $\{y_1, y_2, \dots, y_N\}$ for a given problem, then using a reward model to score and select the highest-scoring candidate:
\begin{equation}
\hat{y} = \arg\max_{y_i \in \{y_1, y_2, \dots, y_N\}} R_\theta(x, y_i)
\end{equation}
\noindent, where $R_\theta(x, y_i)$ represents the reward score for each candidate $y_i$. This technique is especially effective in tasks like mathematical problem-solving and logical inference, where selecting the most plausible solution from a diverse set of outputs improves overall accuracy~\citep{wang2022self}.

\section{Code Preference Model Pretraining}

\subsection{Model Design}
Code Preference Model Pretraining (CodePMP) enhances the sample efficiency of reward models, particularly for reasoning tasks where high-quality preference data is scarce. Traditionally, reward models are finetuned on small, curated datasets, limiting their effectiveness in complex tasks like mathematical reasoning or logical deduction. CodePMP mitigates this limitation by introducing a pretraining phase between basic language model pretraining and finetuning on domain-specific reasoning datasets. This phase leverages a large, diverse dataset of code-preference pairs, enabling the model to learn generalizable patterns and ranking strategies.

CodePMP training involves two components: Reward Modeling (RM) and Language Modeling (LM). In RM, the model is trained on code-preference pairs, learning to assign higher scores to the \textit{chosen} code through a pairwise ranking loss. In LM, only the \textit{chosen} code is used for autoregressive training to maintain the model's general capabilities. The overall loss combines the RM and LM losses, ensuring the model enhances its ranking ability without sacrificing general language modeling performance: $\mathcal{L}_{\text{PMP}} = \mathcal{L}_{\text{RM}} + \mathcal{L}_{\text{LM}}$.

\subsection{Data Construction}
To enable scalable preference model pretraining, we construct a dataset sourced from GitHub, containing over 1.3 billion code files from GitHub repositories. The CodePMP dataset is constructed through a systematic process. First, raw source code is processed by a description summarizer—typically an instruction-tuned CodeLLM—to generate prompts describing the code's functionality. Two CodeLLMs with different capabilities then generate code snippets based on these prompts:
\begin{itemize}
    \item \textbf{Chosen response}: Generated by a more advanced CodeLLM (e.g., 6.7B parameters).
    \item \textbf{Rejected response}: Generated by a less capable CodeLLM (e.g., 1.3B parameters).
\end{itemize}

This process yields pairs of code responses—one chosen and one rejected—which are used for preference modeling. This scalable approach significantly enhances pretraining efficiency, improving performance on downstream tasks. The steps of the CodePMP methodology are outlined systematically in Figure~\ref{fig:codepmp_arch_new}.

\section{Experiments}
In this section, we outline the experimental setup and then the experimental results, highlighting that CodePMP is a highly scalable method.

\begin{table}[t]
\begin{center}
\setlength{\tabcolsep}{3pt}
\begin{tabular}{c|ccc}
\toprule
\textbf{PMP} & \begin{tabular}{c}\textbf{MathShepherd}\\\textbf{-pair}\end{tabular} & \begin{tabular}{c}\textbf{Reclor}\\\textbf{-pair}\end{tabular} & \begin{tabular}{c}\textbf{LogiQA2.0}\\\textbf{-pair}\end{tabular} \\ 
\midrule
\multicolumn{4}{c}{\textbf{Qwen2-1.5B}} \\
\midrule
\ding{55} & 0.7226 & 0.758 & 0.7538 \\ 
\ding{51} & \textbf{0.8186} & \textbf{0.794} & \textbf{0.7774} \\ 
\midrule
\multicolumn{4}{c}{\textbf{Qwen2-7B}} \\
\midrule
\ding{55} & 0.8777 & 0.862 & 0.8263 \\ 
\ding{51} & \textbf{0.9274} & \textbf{0.874} & \textbf{0.8441} \\ 
\bottomrule
\end{tabular}
\caption{Reward model accuracy comparison: CodePMP-initialized models perform better on reasoning test sets, showing better discrimination ability.}
\label{table:codepmp_main_exp_acc}
\end{center}
\end{table}

\subsection{Experimental Settings}
\subsubsection{CodePMP Settings}
\paragraph{Data Construction} We generate code preference pairs by using the deepseek-coder-6.7b-instruct model as the strong CodeLLM to generate \textit{chosen} responses and the deepseek-coder-1.3b-instruct model as the weak CodeLLM to generate \textit{rejected} responses. The constructed CodePMP dataset includes 28 million files and 19 billion tokens. The diverse datasets provide sufficiently broad prompt coverage for preference model pretraining, which is conducive to the generalization of preference models in reasoning tasks. In addition, the average lengths of the \textit{chosen} and \textit{rejected} responses are similar, ensuring that response length does not bias the CodePMP learning process. Details are provided in Appendix.

\paragraph{CodePMP Training} By default, we initialize the preference models with the publicly available Qwen models~\citep{yang2024qwen2}, using different model sizes, specifically Qwen2-1.5B and Qwen2-7B. Detailed hyperparameters for CodePMP training are provided in Appendix.

\subsubsection{Reasoning Finetuning Settings}
We evaluate CodePMP on mathematical and logical reasoning tasks using dedicated preference datasets. For mathematical reasoning, we finetune reward models on MathShepherd-pair dataset, derived from MathShepherd~\citep{wang2023math}, while logical reasoning models use ReClor-pair and LogiQA2.0-pair datasets, derived from ReClor~\citep{yu2020reclor} and LogiQA2.0~\citep{liu2023logiqa} respectively. Each model is finetuned on its corresponding training set and evaluated on its respective holdout test set for accuracy assessment. Implementation details for dataset construction and hyperparameters are provided in Appendix.

\subsubsection{Evaluation Settings}
Following~\citep{zhang2024generative}, we evaluate using two metrics: (1) \textbf{RM Accuracy} measures the reward model's ability to distinguish chosen from rejected solutions on holdout test sets, providing insight into the model's ability to classify individual sequences; and (2) \textbf{Best-of-N (BoN) Accuracy} assesses the percentage of correct solutions selected by the RM from $N$ candidate responses, evaluating the model's group-wise ranking performance and ability to identify the best answer from multiple candidates. We use MetaMath-Mistral-7B~\citep{yu2023metamath} as the generator for BoN evaluation.

We evaluate on GSM8K~\citep{cobbe2021training} and MATH~\citep{hendrycks2021measuring} for mathematical reasoning, and ReClor~\citep{yu2020reclor} and LogiQA2.0~\citep{liu2023logiqa} for logical reasoning. For logical reasoning tasks, we use multiple-choice accuracy (equivalent to Best-of-4) where the RM ranks four manually annotated options, as logical reasoning questions typically consist of paragraphs followed by statements to be judged, making standard BoN evaluation challenging.

\subsection{Experimental Results}

\subsubsection{RM Accuracy Results}
We first compare RM accuracy on the holdout test set with and without CodePMP initialization. As shown in Table~\ref{table:codepmp_main_exp_acc}, RM finetuned with CodePMP initialization achieves higher accuracy on both 1.5B and 7B models across mathematical and logical reasoning tasks, demonstrating that CodePMP enhances the model's ability to differentiate correct from incorrect reasoning. Moreover, CodePMP exhibits strong generalization, yielding significant improvements across different reasoning tasks.

\begin{figure}[t]
\begin{center}
    \begin{subfigure}[b]{0.5\textwidth}
        \centering
        \includegraphics[width=\linewidth]{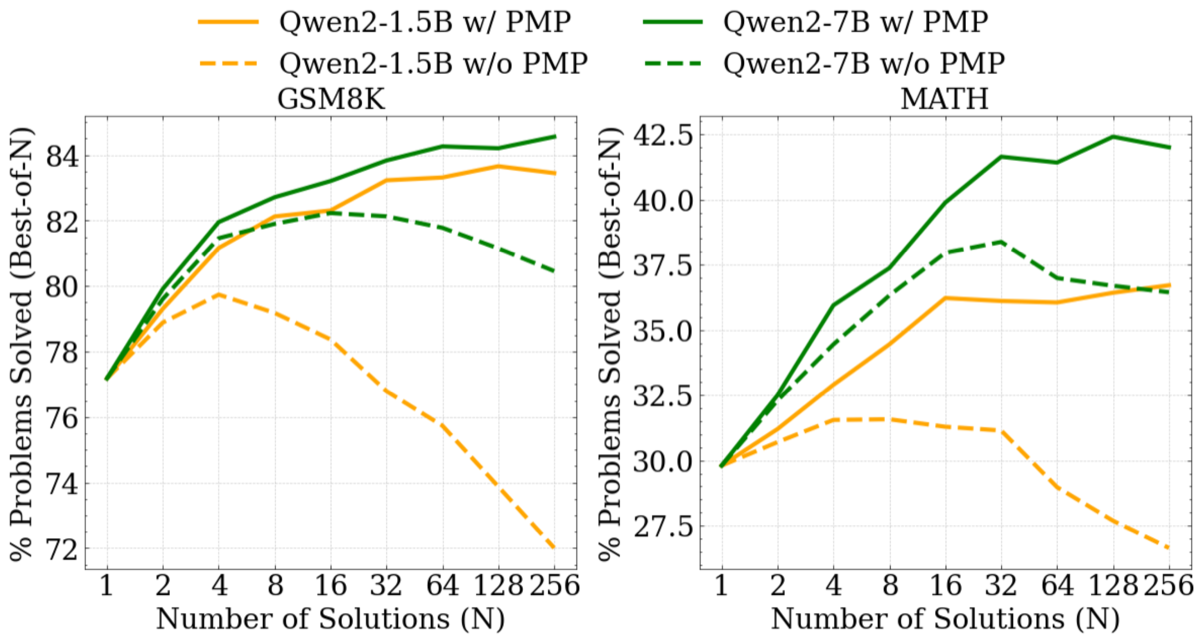}
        \caption{BoN accuracies on mathematical reasoning.}
        \label{fig:codepmp_main_exp_acc_bon}
    \end{subfigure}
    \hfill
    \begin{subfigure}[b]{0.45\textwidth}
        \centering
        \includegraphics[width=\linewidth]{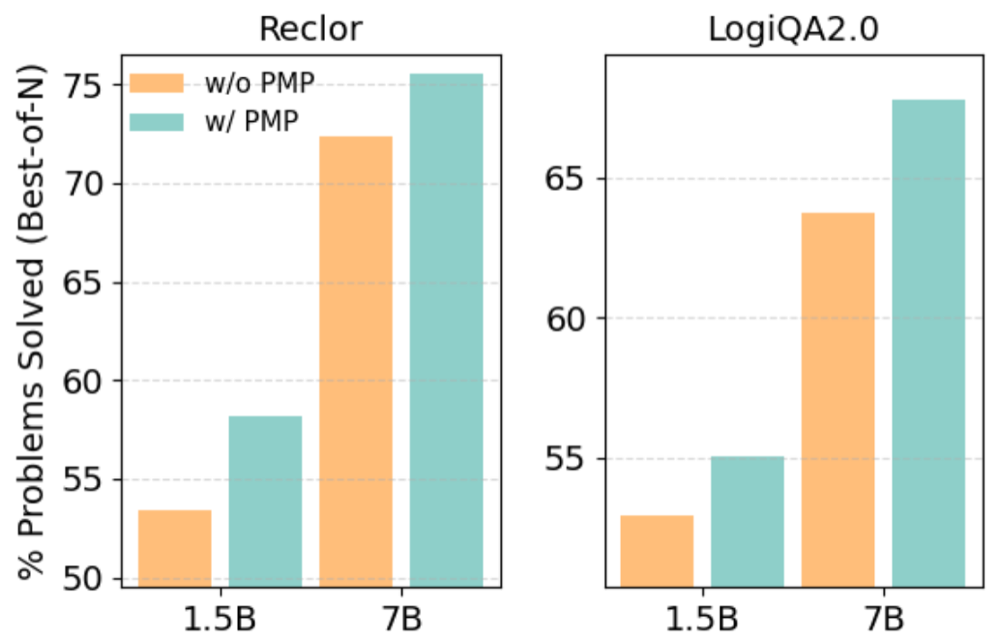}
        \caption{BoN \textit{(N=4)} accuracies on logical reasoning.}
        \label{fig:main_exp_bon_logic}
    \end{subfigure}
    \caption{Best-of-N accuracy comparison: CodePMP-initialized models outperform baselines across various N values, showing superior ranking capabilities.}
    \label{fig:combined_main_exp_bon}
\end{center}
\end{figure}

\subsubsection{BoN Accuracy Results}
Evaluations across reasoning tasks demonstrate that CodePMP-initialized RMs consistently achieve higher BoN accuracy on both mathematical and logical reasoning tasks for all model sizes (Figure~\ref{fig:combined_main_exp_bon}). CodePMP models maintain performance advantages even as N increases to 256, while non-CodePMP models exhibit significant accuracy degradation at higher N values, highlighting CodePMP's stability.

This aligns with research on BoN sampling~\citep{chow2024inferenceawarefinetuning} that identifies an inflection point where performance typically deteriorates beyond certain N thresholds due to increased base policy stochasticity and verifier misalignment. CodePMP-initialized models demonstrate greater stability at higher N values, suggesting improved alignment with true reward signals and enhanced robustness to noise amplification inherent in large-N sampling.

For logical reasoning, the performance gap appears smaller as testing was limited to N=4, while mathematical reasoning extended to N=256, suggesting potential for amplified advantages in logical reasoning with increased N values.

\subsubsection{Sample Efficiency Analysis}
\begin{figure}[t]
\vspace{-4pt}
\begin{center}
    \begin{subfigure}[b]{0.48\columnwidth}
        \centering
        \includegraphics[width=\linewidth]{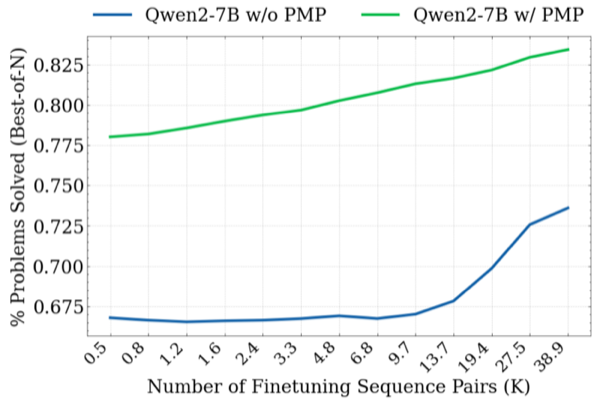}
        \caption{GSM8K / \textcolor{red}{7B}}
    \end{subfigure}
    \hfill
    \begin{subfigure}[b]{0.48\columnwidth}
        \centering
        \includegraphics[width=\linewidth]{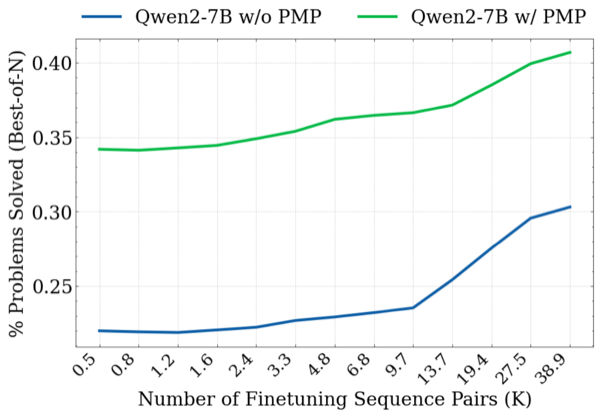}
        \caption{MATH / \textcolor{red}{7B}}
    \end{subfigure}
    
    \vspace{0.3em}
    
    \begin{subfigure}[b]{0.48\columnwidth}
        \centering
        \includegraphics[width=\linewidth]{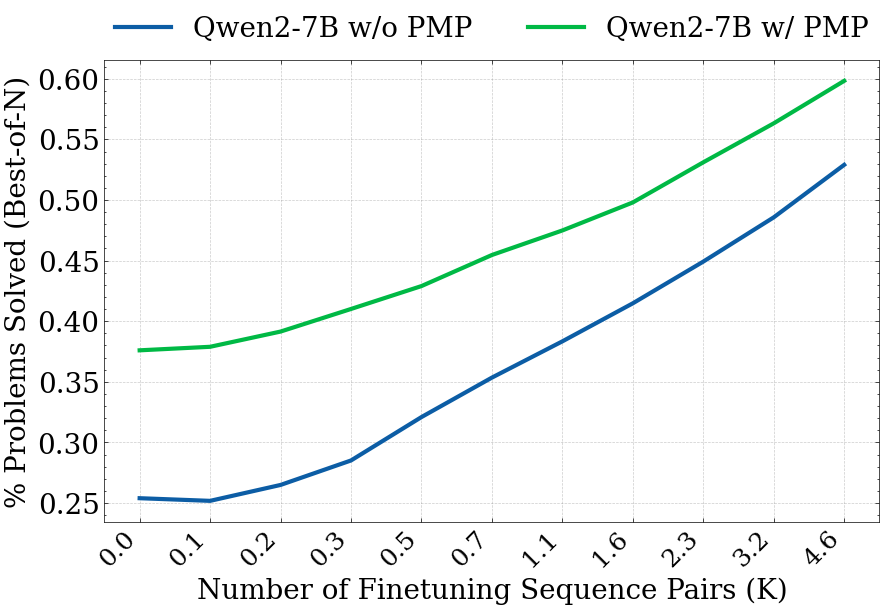}
        \caption{Reclor / \textcolor{red}{7B}}
    \end{subfigure}
    \hfill
    \begin{subfigure}[b]{0.48\columnwidth}
        \centering
        \includegraphics[width=\linewidth]{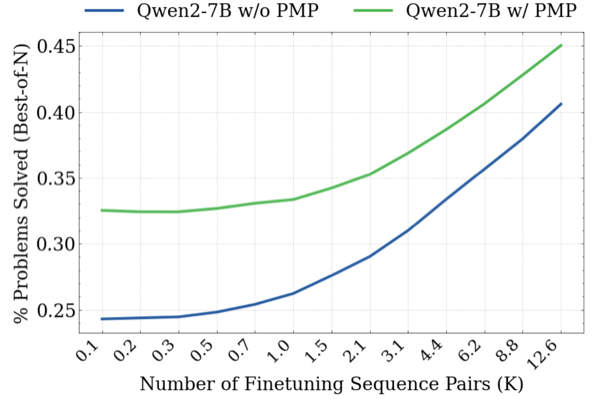}
        \caption{LogiQA2.0 / \textcolor{red}{7B}}
    \end{subfigure}
    \caption{Sample efficiency comparison for 7B models: CodePMP-initialized reward models achieve higher Best-of-N accuracy with the equivalent sample sizes, showing better data efficiency. Horizontal axis scales by $\sqrt{2}$. Green: with CodePMP; Blue: without CodePMP.}
    \label{fig:rm_bon_scaling}
\end{center}
\end{figure}

To assess CodePMP's impact on sample efficiency, we evaluated models with varying finetuning dataset sizes following best practices~\citep{kaplan2020scaling}. Figure~\ref{fig:rm_bon_scaling} shows that CodePMP-initialized models consistently outperform baselines across all dataset sizes, with CodePMP achieving with just 0.5k samples what baseline models require 40k samples to match—an 80× efficiency improvement. This advantage, while diminishing with larger datasets, significantly reduces annotation costs for developing effective reward models.

\subsubsection{Scalability Analysis}
\begin{figure}[t]
    \centering
    \begin{subfigure}[b]{0.48\columnwidth}
        \centering
        \includegraphics[width=\linewidth]{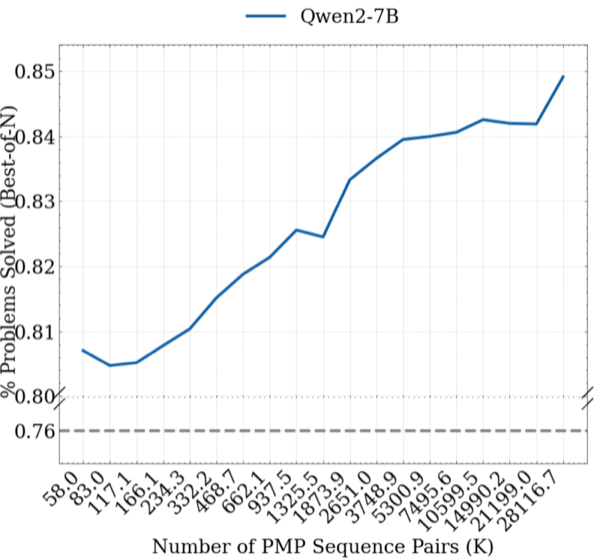}
        \caption{GSM8K / \textcolor{red}{7B}}
    \end{subfigure}
    \hfill
    \begin{subfigure}[b]{0.48\columnwidth}
        \centering
        \includegraphics[width=\linewidth]{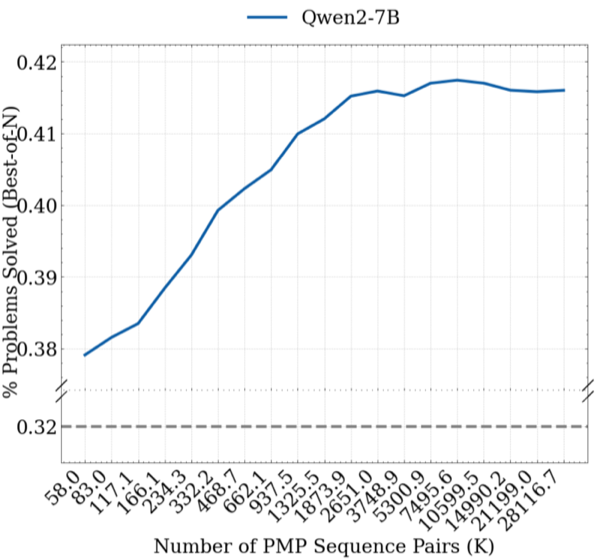}
        \caption{MATH / \textcolor{red}{7B}}
    \end{subfigure}
    
    \vspace{0.3em}
    
    \begin{subfigure}[b]{0.48\columnwidth}
        \centering
        \includegraphics[width=\linewidth]{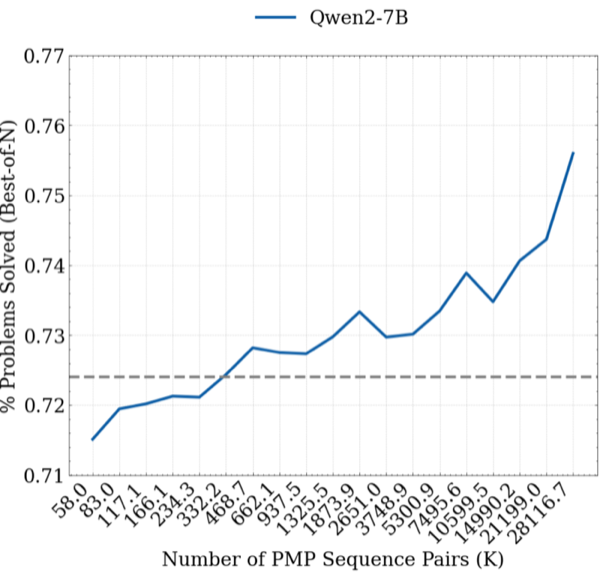}
        \caption{Reclor / \textcolor{red}{7B}}
    \end{subfigure}
    \hfill
    \begin{subfigure}[b]{0.48\columnwidth}
        \centering
        \includegraphics[width=\linewidth]{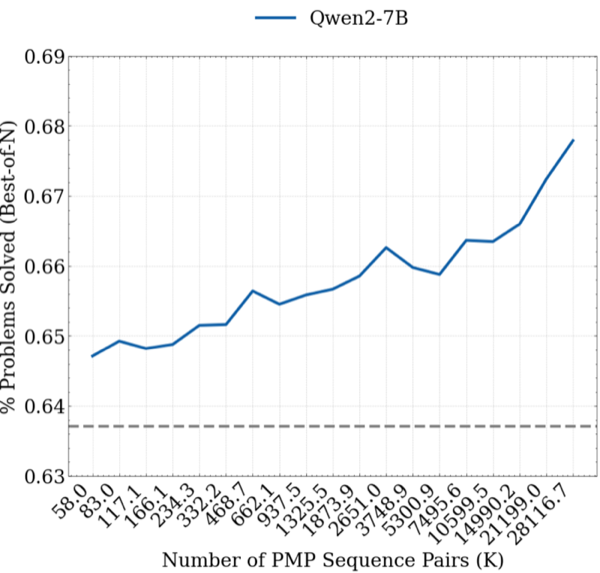}
        \caption{LogiQA2.0 / \textcolor{red}{7B}}
    \end{subfigure}
    \caption{Scaling analysis of CodePMP for 7B models: more code-preference pairs consistently improve Best-of-N accuracy across reasoning tasks without diminishing returns. Horizontal axis scales by $\sqrt{2}$; gray dashed lines show baseline performance without CodePMP.}
    \label{fig:codepmp_exp_pmp_scaling}
\end{figure}

A key benefit of using code data for PMP is the vast availability of publicly accessible, high-quality code-preference pairs, ensuring diversity. To validate scalability, we vary the number of training pairs for CodePMP and retrain models with different amounts of data. As shown in Figure~\ref{fig:codepmp_exp_pmp_scaling}, increasing the number of code-preference pairs consistently improves BoN accuracy in both mathematical and logical reasoning tasks across model sizes, with no sign of diminishing returns. This indicates that further scaling the code-preference data would likely yield additional performance gains, underscoring the importance of building a scalable PMP pipeline.

\section{Ablation Studies}
This section presents a detailed analysis of CodePMP design. Unless otherwise stated, all experiments use the 1B model due to resource limitations and present the results of mathematical reasoning due to page limitation. More ablation studies refer to Appendix.

\subsection{Impact of Pair Construction}
\begin{figure}[t]
    \centering
    \begin{subfigure}[b]{0.49\textwidth}
        \centering
        \includegraphics[width=\linewidth]{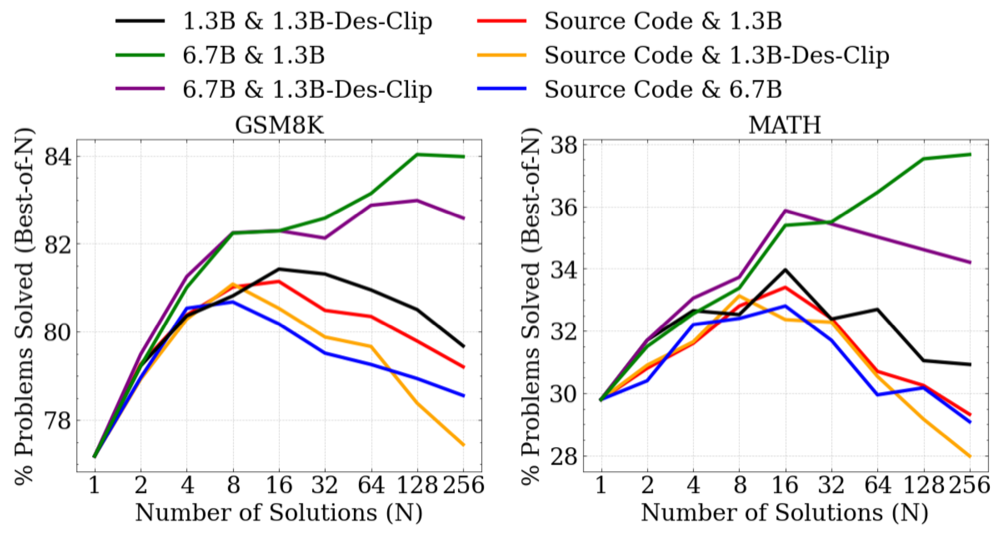}
        \caption{Different construction methods.}
        \label{fig:res_cmp_pair_construction}
    \end{subfigure}
    \hfill
    \begin{subfigure}[b]{0.49\textwidth}
        \centering
        \includegraphics[width=\linewidth]{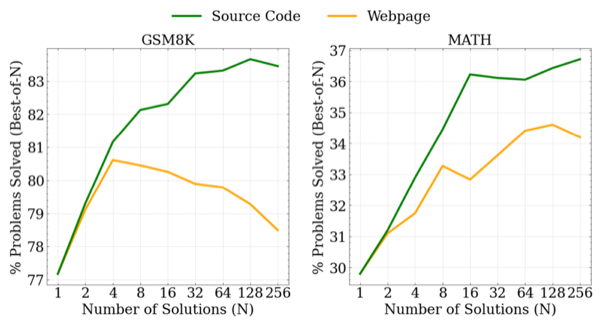}
        \caption{Different pair sources.}
        \label{fig:res_cmp_code_web_pair}
    \end{subfigure}
    \caption{Comparison of BoN accuracy across construction methods and data sources, demonstrating benefits of model-based construction and GitHub code.}
    \label{fig:combined_results}
\end{figure}

\paragraph{GitHub-Sourced Pairs vs Web-Crawled} 
We compare GitHub-sourced code with web-crawled data~\citep{askell2021general} from platforms such as StackExchange and Reddit. As shown in Figure~\ref{fig:res_cmp_code_web_pair}, GitHub-sourced pairs (``Source Code'') consistently outperform those from web platforms (``Webpage''), particularly as the number of solutions (N) increases. Moreover, the performance improvement of GitHub-sourced pairs shows no sign of plateauing, highlighting the importance of diverse, high-quality source code in building a scalable PMP pipeline.

\paragraph{Model Generated Data vs Human Data} 
We compare various pair construction methods generated by different models. In Figure~\ref{fig:res_cmp_pair_construction}, the samples before the ``\&'' are positive, and those after are negative. ``Source Code'' refers to the original code snippet, while ``1.3B-Des-Clip'' indicates that 10\% of the code description is removed before being input into a 1.3B CodeLLM to generate a rejected response. The green lines represent CodePMP's choice. Results show that pairing positive samples from the 7B model with negative samples from the 1.5B model consistently delivers the best performance across all test sets. Given that code execution can generate reliable outputs, future work will explore incorporating execution feedback to create more accurate preference pairs.

\begin{figure}[t]
\vspace{-4pt}
    \centering
    \begin{subfigure}[b]{0.5\textwidth}
        \centering
        \includegraphics[width=\linewidth]{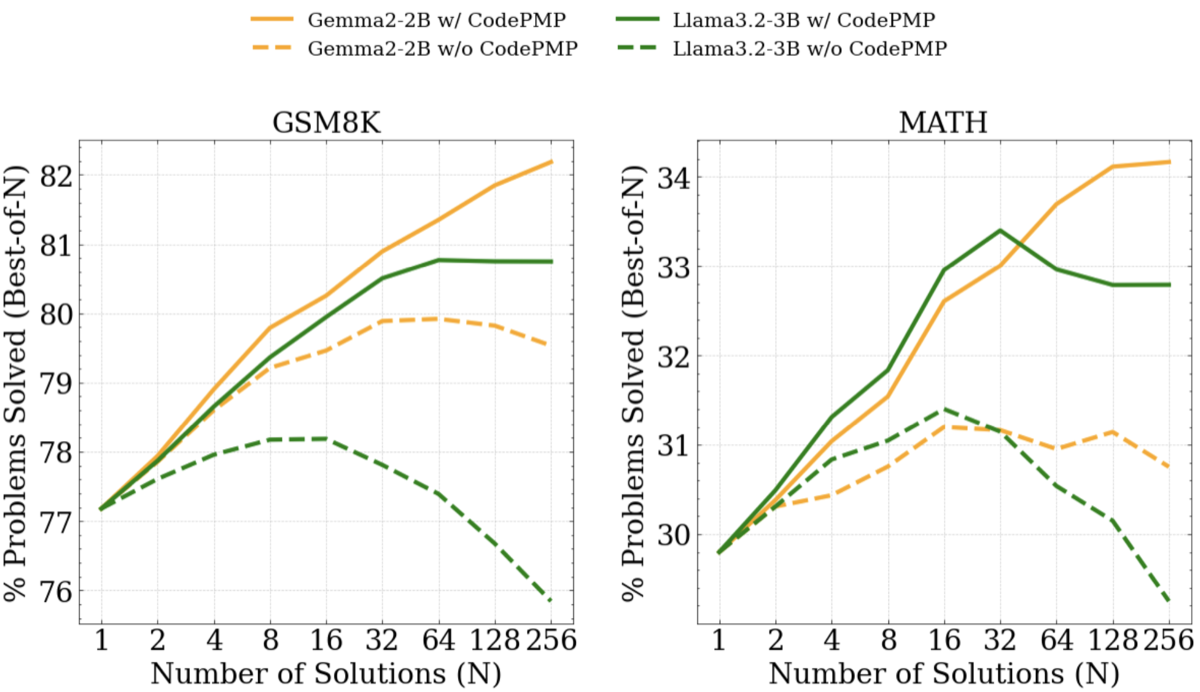}
        \caption{BoN accuracies on mathematical reasoning.}
        \label{fig:gemma&llama}
    \end{subfigure}
    \hfill
    \begin{subfigure}[b]{0.45\textwidth}
        \centering
        \includegraphics[width=\linewidth]{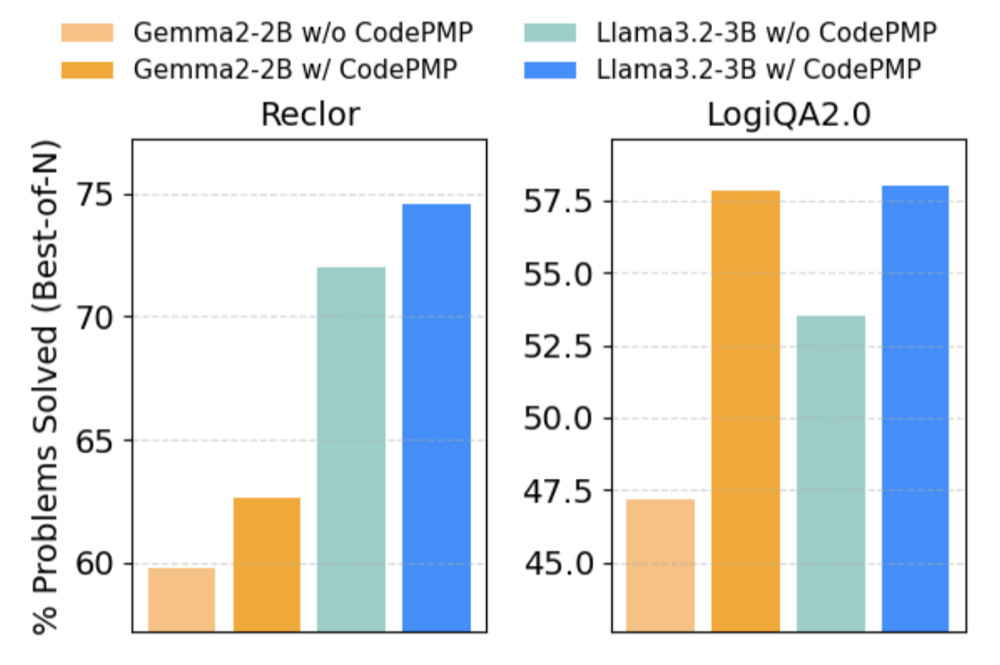}
        \caption{BoN \textit{(N=4)} accuracies on logical reasoning.}
        \label{fig:gemmakllama_ligical}
    \end{subfigure}
    \caption{Cross-architecture performance comparison: CodePMP enhances reasoning performance across different model families (Gemma2 and Llama3.2), showing broad applicability.}
    \label{fig:family}
\end{figure}

\subsection{Impact of Loss Function}
\label{sec:ablation_loss}
CodePMP integrates both Reward Modeling (RM) and Language Modeling (LM) loss components. To evaluate their contributions, we conducted experiments comparing three configurations: RM loss only, LM loss only, and the combined approach. As shown in Table~\ref{tab:loss_comparison}, the combined loss function consistently outperforms single-loss variants across all Best-of-N evaluation settings, with particularly notable improvements on the challenging MATH dataset. This empirical evidence indicates a complementary relationship where RM loss enhances preference ranking while LM loss preserves general language capabilities, collectively yielding more robust reward model performance.

\begin{table}[t]
    \begin{center}
    \setlength{\tabcolsep}{3pt}
    \begin{tabular}{c|ccc}
    \toprule
    \textbf{BoN} & \textbf{RM Loss} & \textbf{LM Loss} & \textbf{RM + LM Loss} \\
    \midrule
    \multicolumn{4}{c}{\textbf{GSM8K}} \\
    \midrule
    N=32  & 0.834   & 0.8317   & \textbf{0.8393}  \\ 
    N=64  & 0.8362  & 0.8271   & \textbf{0.8453}  \\ 
    N=128 & 0.8332  & 0.8309   & \textbf{0.8362}  \\ 
    N=256 & 0.8271  & 0.8226   & \textbf{0.8484}  \\
    \midrule
    \multicolumn{4}{c}{\textbf{MATH}} \\
    \midrule
    N=32  & 0.344   & 0.376    & \textbf{0.418}  \\ 
    N=64  & 0.358   & 0.376    & \textbf{0.424}  \\ 
    N=128 & 0.366   & 0.354    & \textbf{0.434}  \\ 
    N=256 & 0.362   & 0.372    & \textbf{0.41}   \\
    \bottomrule
    \end{tabular}
    \caption{Loss function comparison.}
    \label{tab:loss_comparison}
    \end{center}
\end{table}

\subsection{Cross-Architecture Generalization}
\label{sec:ablation_different_family}

To assess CodePMP's generalization capabilities beyond the Qwen architecture family, we evaluated its effectiveness with Gemma2 and Llama3.2 as PMP/RM backbones on GSM8K, MATH, Reclor, and LogiQA-v2 benchmarks. As shown in Figure~\ref{fig:family}, CodePMP:
(1) Consistently enhances reasoning performance across all model families, and 
(2) Improves robustness at larger $N$ values, mitigating performance degradation observed in non-initialized models.These results demonstrate that CodePMP generalizes effectively across diverse model architectures, suggesting broad applicability of the approach.

\subsection{Performance on Larger Backbone Model}
\label{sec:ablation_larger_backbone}

To investigate CodePMP's performance on larger model scales, we applied the technique to Qwen2-72B. Table~\ref{tab:qwen2_72b_combined} presents results across mathematical and logical reasoning tasks.

\begin{table}[!t]
    \centering
    \setlength{\tabcolsep}{3pt}
    \begin{tabular}{c|cc|cc}
    \toprule
    \multirow{2}{*}{\textbf{BoN}} & \textbf{w/o PMP} & \textbf{w/ PMP} & \textbf{w/o PMP} & \textbf{w/ PMP} \\
    & \multicolumn{2}{c|}{\textbf{GSM8K}} & \multicolumn{2}{c}{\textbf{MATH}} \\
    \midrule
    \textbf{N=1}   & 0.7718 & 0.7718 & 0.298 & 0.298 \\
    N=4   & 0.8453 & 0.8453 & 0.424 & 0.424 \\
    N=32  & 0.8529 & \textbf{0.8628} & 0.488 & \textbf{0.500} \\
    N=256 & 0.8249 & \textbf{0.8400} & 0.506 & \textbf{0.514} \\
    \midrule
    \textbf{BoN} & \multicolumn{2}{c|}{\textbf{Reclor}} & \multicolumn{2}{c}{\textbf{LogiQA-v2}} \\
    \midrule
    N=4   & 0.894 & \textbf{0.918} & 0.7117 & \textbf{0.7927} \\
    \bottomrule
    \end{tabular}
    \caption{Performance comparison on reasoning tasks for Qwen2-72B with and without CodePMP initialization. Note that only $N=4$ was tested for Reclor and LogiQA-v2.}
    \label{tab:qwen2_72b_combined}
\end{table}

Results show consistent improvements with CodePMP initialization across all benchmarks. Notably, performance gains increase with larger $N$ values on challenging tasks like MATH, indicating that CodePMP's benefits scale effectively to larger model architectures. The significant improvement on logical reasoning tasks further demonstrates CodePMP's scalability and broad applicability.

\subsection{Performance on More Powerful Generator}
\label{sec:ablation_powerful_generator}

To determine whether CodePMP maintains its effectiveness with more sophisticated generators, we conducted experiments with two advanced models: Qwen2-Math-7B-Instruct (specialized for mathematical reasoning) and Qwen2.5-32B-Instruct (a substantially larger general-purpose model).

\begin{table}[!t]
    \centering
    \setlength{\tabcolsep}{3pt}
    \begin{tabular}{c|cc|cc}
    \toprule
    \multirow{2}{*}{\textbf{BoN}} & \multicolumn{2}{c|}{\textbf{GSM8K}} & \multicolumn{2}{c}{\textbf{MATH}} \\
    & \textbf{w/o PMP} & \textbf{w/ PMP} & \textbf{w/o PMP} & \textbf{w/ PMP} \\
    \midrule
    N=4   & 0.8544 & \textbf{0.8931} & 0.690 & \textbf{0.724} \\
    N=32  & 0.8446 & \textbf{0.8795} & 0.643 & \textbf{0.698} \\
    N=256 & 0.8256 & \textbf{0.8590} & 0.614 & \textbf{0.690} \\
    \bottomrule
    \end{tabular}
    \caption{BoN accuracy with specialized mathematical generator (Qwen2-Math-7B-Instruct).}
    \label{tab:math_generators}
\end{table}

\begin{table}[!t]
    \centering
    \setlength{\tabcolsep}{3pt}
    \begin{tabular}{c|cc|cc}
    \toprule
    \multirow{2}{*}{\textbf{BoN}} & \multicolumn{2}{c|}{\textbf{GSM8K}} & \multicolumn{2}{c}{\textbf{MATH}} \\
    & \textbf{w/o PMP} & \textbf{w/ PMP} & \textbf{w/o PMP} & \textbf{w/ PMP} \\
    \midrule
    N=4   & 0.9604 & \textbf{0.9688} & 0.798 & \textbf{0.820} \\
    N=32  & 0.9573 & \textbf{0.9581} & 0.768 & \textbf{0.792} \\
    N=256 & 0.9566 & \textbf{0.9634} & 0.752 & \textbf{0.798} \\
    \bottomrule
    \end{tabular}
    \caption{BoN accuracy with large-scale generator (Qwen2.5-32B-Instruct).}
    \label{tab:large_generators}
\end{table}

Tables~\ref{tab:math_generators} and~\ref{tab:large_generators} demonstrate that CodePMP's benefits persist across different generator architectures. With the specialized Qwen2-Math-7B-Instruct (Table~\ref{tab:math_generators}), we observe substantial improvements on both GSM8K and MATH. These gains remain consistent with the much larger Qwen2.5-32B-Instruct model (Table~\ref{tab:large_generators}), despite it being significantly larger than both the preference pair generation models (less than 7B parameters) and the reward model itself (Qwen2-7B).

These findings demonstrate that reward models trained on synthetic preference data from smaller models can effectively guide more powerful and specialized generators, confirming CodePMP's robustness and cross-scale applicability. This is particularly significant as it suggests that relatively modest investments in reward model training can yield benefits even when deployed with state-of-the-art generation systems.

\subsection{Performance on General RM Benchmarks}
\label{sec:ablation_general_rm}

We further evaluate CodePMP on general reward modeling benchmarks (RMBench) to assess its applicability beyond reasoning tasks. RMBench provides an out-of-domain assessment covering various tasks including summarization, chat quality, and safety. As shown in Table \ref{table:rm_bench_results}, models fine-tuned with PMP consistently outperform those without PMP across various model sizes and tasks.

\begin{table*}[!t]
\begin{center}
\setlength{\tabcolsep}{6pt}
\begin{tabular}{cc|ccccc}
\toprule
\multirow{2}{*}{\textbf{Model}} & \multirow{2}{*}{\textbf{PMP}} & \multicolumn{5}{c}{\textbf{RMBench}} \\
 & & \textbf{Summary} & \textbf{Chat} & \textbf{Chat Hard} & \textbf{Safety} & \textbf{Reasoning} \\ \midrule
\multirow{2}{*}{1.5B} & \ding{55} & 0.4154 & 0.4804 & \textbf{0.5351} & 0.3665 & 0.2751 \\ \cline{2-7} \addlinespace[3pt]
 & \ding{51} & \textbf{0.6126} & \textbf{0.9050} & 0.4364 & \textbf{0.3698} & \textbf{0.6041} \\ \midrule
\multirow{2}{*}{7B} & \ding{55} & 0.5839 & 0.4972 & 0.5022 & \textbf{0.5240} & 0.6804 \\ \cline{2-7} \addlinespace[3pt]
&  \ding{51} & \textbf{0.7668} & \textbf{0.9413} & \textbf{0.5373} & 0.4906 & \textbf{0.9116} \\ \bottomrule
\end{tabular}
\caption{Performance on RMBench shows that CodePMP generalizes well across various general LLM tasks.}
\label{table:rm_bench_results}
\end{center}
\end{table*}

These results demonstrate that CodePMP enhances performance not only in reasoning and coding tasks but also improves generalization across a broad range of RM benchmarks. These findings provide compelling evidence for CodePMP's broad applicability across multiple domains beyond the reasoning tasks that were our primary focus.

\section{Related Works}
\paragraph{Reward Modeling} 
In the context of RLHF, reward models (RMs) have traditionally employed ranking models like Bradley-Terry and Plackett-Luce to represent human preferences~\citep{bradley1952rank, plackett1975, cobbe2021training, self-critique-models, lightman2023let, wang2023math, uesato2022solving, luo2024improve, yu2024ovm, stiennon2020learning, nakano2021webgpt}. More recently, probability-based approaches have emerged, offering more precise predictions. Additionally, models such as Critique-out-Loud~\citep{ankner2024} enhance RMs by integrating natural language feedback. Generative reward models (GRMs) further boost sample efficiency. Preference Modeling Pretraining (PMP)~\citep{askell2021general} introduces a novel pretraining phase, utilizing large-scale pairwise ranking data to enhance RM performance. Despite these advancements, many methods are hindered by the reliance on expensive manual annotations or limited datasets, constraining scalability. CodePMP mitigates this by automating preference data generation from code, significantly improving RM sample efficiency and reducing dependency on manual data collection.

\paragraph{Code Training} 
The inclusion of code in LLM pretraining has led to marked improvements in tasks such as commonsense reasoning~\citep{madaan2022language} and mathematical problem-solving~\citep{liang2022holistic, shao2024deepseekmath, yang2024qwen2}. Furthermore, code enhances general reasoning capabilities~\citep{muennighoff2023scaling, fu2022gptroadmap, ma2023training}. Recent studies~\citep{dong2023abilities, ma2023training} indicate that incorporating code during supervised finetuning strengthens LLMs, particularly in complex decision-making tasks. CodePMP takes a pioneering approach by utilizing scalable, synthetically generated code preference pairs, reducing the dependence on manual annotation~\citep{dubey2024llama, geminiteam2024gemini, groeneveld2024olmo, bi2024deepseek}. This methodology enhances sample efficiency and scalability in reasoning-intensive tasks, presenting new opportunities for further improving LLM performance.

\paragraph{LLM Reasoning}
Improving reasoning capabilities in LLMs remains a significant challenge, with various advanced methods being proposed. Chain of Thought (CoT) prompting~\citep{Wei2022ChainOfThought, Fu2023ComplexCoT} improves reasoning by generating intermediate steps, while CoT combined with supervised finetuning (SFT) further enhances performance~\citep{cobbe2021training, Liu2024MMIQC, yu2023metamath}. Other approaches focus on expanding inference time computation, such as problem decomposition~\citep{zhou2022least}, search-based methods like MCTS~\citep{xu2023no}, and using LLMs as verifiers~\citep{Huang2022SelfImprove, Luo2023bCritique}. Reward models, including outcome-based (ORM) and process-based (PRM), have also shown success, with PRM delivering superior results~\citep{lightman2023let, wang2023math}. Encouragingly, CodePMP introduces a scalable preference model pretraining phase that can integrate seamlessly with all the aforementioned techniques.

\section{Conclusion and Future Works}
We propose \textbf{CodePMP}, a scalable preference model pretraining method that leverages synthetic code-preference pairs to boost reasoning in large language models. Experiments demonstrate that CodePMP markedly enhances both sample efficiency and performance across diverse reasoning tasks, validating the effectiveness of code-based preference pretraining. Future directions include CodePrMP, which will utilize compiler/interpreter verifiability for low-cost process supervision, and GenPMP, aimed at improving generative reward models through code-based pretraining.

\section*{Limitations}
Our current implementation has several limitations. First, the synthetic preference pairs rely on models with predetermined parameter sizes, potentially missing nuanced preference signals that more sophisticated approaches might capture. While we demonstrate broad applicability across model families, architectural differences may affect performance in ways not fully explored in this work. Our reliance on GitHub data introduces potential biases stemming from the composition of public repositories. Additionally, our evaluation focuses primarily on mathematical and logical reasoning, leaving the method's effectiveness for other reasoning modalities (e.g., commonsense or causal reasoning) less thoroughly examined. Future work should address these limitations to further enhance the generalizability and robustness of the approach.

\section*{Ethics Statement}
CodePMP introduces several important ethical considerations. By enhancing LLMs' reasoning capabilities, it could significantly impact decision-making systems that affect human lives, necessitating careful deployment and monitoring. While we utilize publicly available code, we recognize the importance of intellectual property rights and have focused on data with permissive licenses. Our approach reduces reliance on human annotation, potentially mitigating certain biases while possibly introducing others derived from the training data or model preferences. These trade-offs require ongoing evaluation and refinement to ensure fair and beneficial applications. As with any technology that enhances AI capabilities, responsible deployment with appropriate safeguards is essential.

\bibliography{custom}

\appendix
\section{Hyperparameters}\label{apendix:hyperparameters}
We outline key hyperparameters in this section. In the tables, WSD refers to the warmup-stable-decay learning rate scheduler~\citep{hu2024minicpm}, which has the benefit of reducing the time required for scaling law experiments. Specifically, Table~\ref{codepmp_hyperparameters} lists the hyperparameters for CodePMP training, Table~\ref{codepmp_math_ft} details those for mathematical reasoning RM fine-tuning, Table~\ref{codepmp_logic_ft} covers logical reasoning RM fine-tuning, and Table~\ref{codepmp_bon_gen} presents the hyperparameters for BON generation.

\begin{table}[!t]
\begin{tabular}{lcc}
\hline
\textbf{Hyperparameter} & \textbf{Qwen2-1.5B} & \textbf{Qwen2-7B} \\
\hline
epoch & 1 & 1 \\
batch size & 1024 & 1024 \\
learning rate & 3e-6 & 1e-6 \\
lr scheduler & WSD & WSD \\
warmup ratio & 0.03 & 0.03 \\
decay ratio & 0.1 & 0.1 \\
weight decay & 0.1 & 0.1 \\
max length & 1024 & 1024 \\
\hline
\end{tabular}
\caption{CodePMP training hyperparameters.}
\label{codepmp_hyperparameters}
\end{table}

\begin{table}[!t]
\begin{tabular}{lcc}
\hline
\textbf{Hyperparameter} & \textbf{Qwen2-1.5B} & \textbf{Qwen2-7B} \\
\hline
epoch & 1 & 1 \\
batch size & 64 & 64 \\
learning rate & 1e-6 & 3e-7 \\
lr scheduler & WCD & WCD \\
warmup ratio & 0.03 & 0.03 \\
weight decay & 0 & 0 \\
max length & 1024 & 1024 \\
\hline
\end{tabular}
\caption{Mathematical reasoning RM finetuning hyperparameters.}
\label{codepmp_math_ft}
\end{table}

\begin{table}[!t]
\begin{tabular}{lcc}
\hline
\textbf{Hyperparameter} & \textbf{Qwen2-1.5B} & \textbf{Qwen2-7B} \\
\hline
epoch & 1 & 1 \\
batch size & 64 & 64 \\
learning rate & 1e-5 & 1e-5 \\
lr scheduler & WCD & WCD \\
warmup ratio & 0.25 & 0.25 \\
weight decay & 0 & 0 \\
max length & 1024 & 1024 \\
\hline
\end{tabular}
\caption{Logical reasoning RM finetuning hyperparameters.}
\label{codepmp_logic_ft}
\end{table}

\begin{table}[!t]
\begin{tabular}{lcc}
\hline
\textbf{Hyperparameter} & \textbf{MetaMath-Mistral-7B} \\
\hline
temperature & 0.7 \\
top-p & 1 \\
\hline
\end{tabular}
\caption{Best-of-N generation hyperparameters.}
\label{codepmp_bon_gen}
\end{table}

\section{CodePMP Dataset Statistics} \label{appendix_codepmp_data_stats}
Table~\ref{codepmp_pair_stats} presents the average token lengths of responses in the CodePMP dataset. The similar lengths between chosen and rejected responses (194.5 vs. 189.9 tokens) ensure that response length does not introduce bias in the learning process. The dataset comprises 28 million files totaling 19 billion tokens, with Python (13.1B tokens), Jupyter Notebooks (2.1B tokens), and other languages (3.8B tokens) providing diverse coverage that facilitates model generalization.

\begin{table}[!t]
\centering
\begin{tabular}{lcc}
\hline
\textbf{Language} & \textbf{Chosen} & \textbf{Rejected} \\
\hline
Python & 170.0 & 167.0 \\
Notebook & 158.0 & 155.5 \\
Other Languages & 213.2 & 210.0 \\
\hline
Total & \textbf{194.5} & \textbf{189.9} \\
\hline
\end{tabular}
\caption{Average token lengths of responses in the CodePMP dataset by language category.}
\label{codepmp_pair_stats}
\end{table}

\begin{figure*}[htbp]
    \centering
    \begin{subfigure}[b]{0.24\textwidth}
        \centering
        \includegraphics[width=\linewidth]{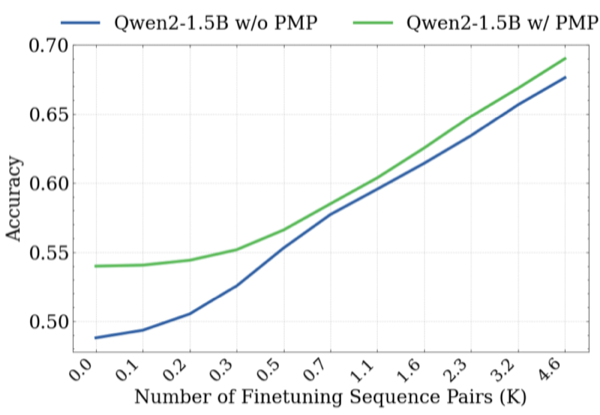}
        \caption{MATH-shepherd / 1.5B}
    \end{subfigure}
    \hfill
    \begin{subfigure}[b]{0.24\textwidth}
        \centering
        \includegraphics[width=\linewidth]{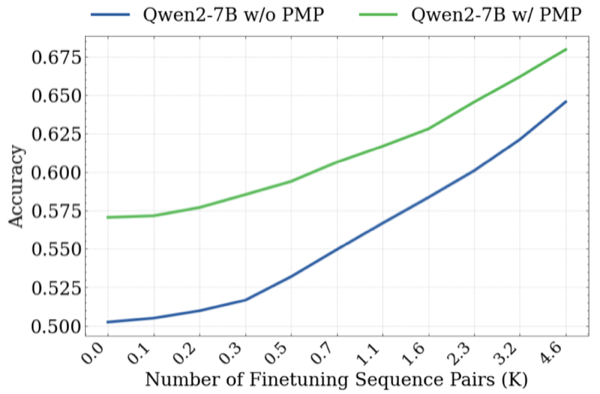}
        \caption{MATH-shepherd / 7B}
    \end{subfigure}
    \hfill
    \begin{subfigure}[b]{0.24\textwidth}
        \centering
        \includegraphics[width=\linewidth]{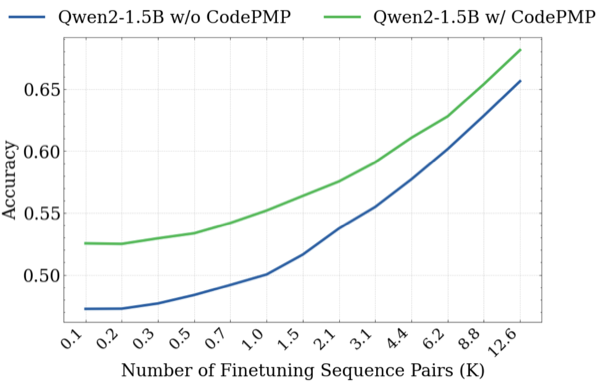}
        \caption{Reclor+Logiqa / 1.5B}
    \end{subfigure}
    \hfill
    \begin{subfigure}[b]{0.24\textwidth}
        \centering
        \includegraphics[width=\linewidth]{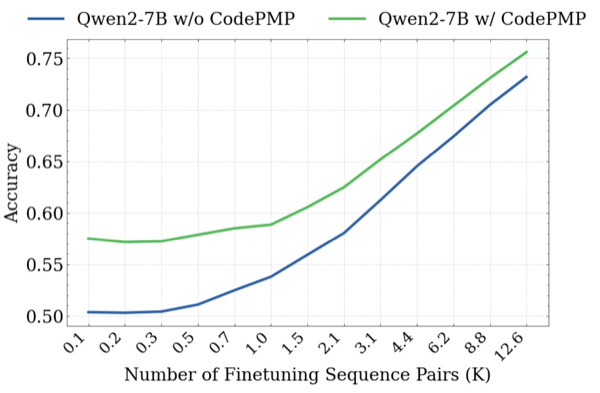}
        \caption{Reclor+Logiqa / 7B}
    \end{subfigure}
    \caption{Comparison of sample efficiency of RM fine-tuning: Trends of RM accuracy with sample size increases.}
    \label{rm_acc_reclor_or_logiqa}
\end{figure*}

\begin{figure*}[htbp]
    \centering
    \begin{subfigure}[b]{0.24\textwidth}
        \centering
        \includegraphics[width=\linewidth]{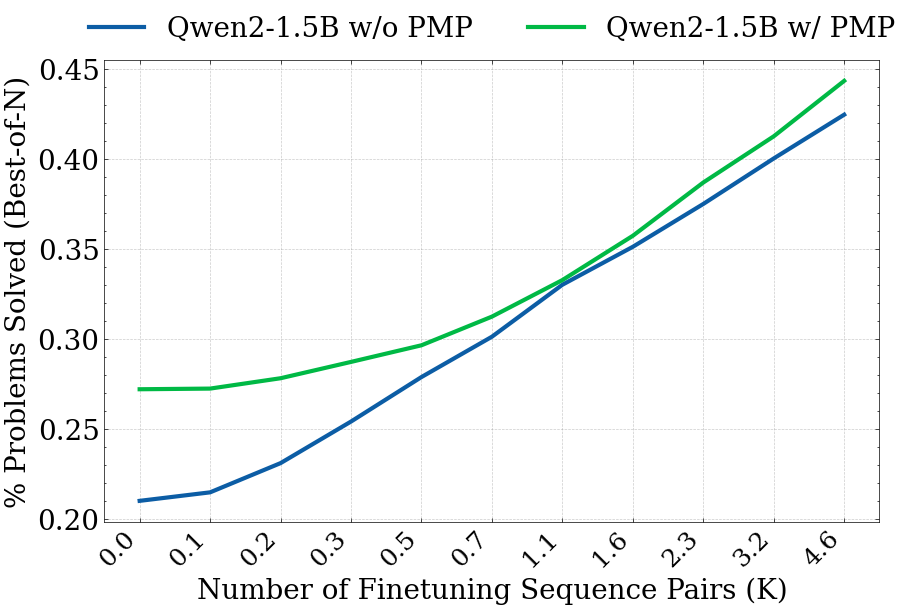}
        \caption{MATH-shepherd / 1.5B}
    \end{subfigure}
    \hfill
    \begin{subfigure}[b]{0.24\textwidth}
        \centering
        \includegraphics[width=\linewidth]{images/rm_bon_reclor_2.png}
        \caption{MATH-shepherd / 7B}
    \end{subfigure}
    \hfill
    \begin{subfigure}[b]{0.24\textwidth}
        \centering
        \includegraphics[width=\linewidth]{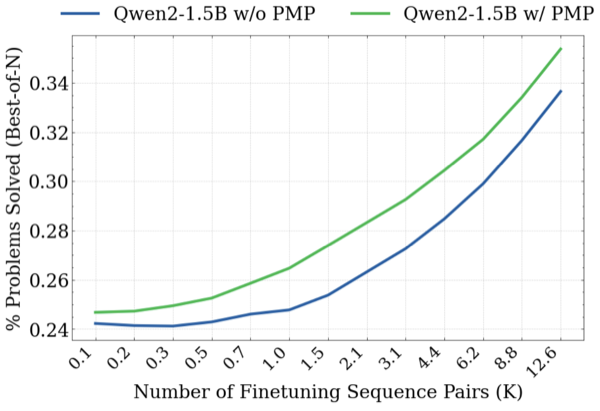}
        \caption{Reclor+Logiqa / 1.5B}
    \end{subfigure}
    \hfill
    \begin{subfigure}[b]{0.24\textwidth}
        \centering
        \includegraphics[width=\linewidth]{images/rm_bon_logiqa_2.png}
        \caption{Reclor+Logiqa / 7B}
    \end{subfigure}
    \caption{Comparison of sample efficiency of RM fine-tuning: Trends of Multi-choice accuracy or Best-of-4 with sample size increases.}
    \label{rm_bon_reclor_or_logiqa}
\end{figure*}

\section{Further Comparisons and Cross-Domain Evaluations}
\label{sec:further_comparisons}

\subsection{Comparison with Majority Voting}
\label{sec:majority_voting}

We compare CodePMP with a majority-voting baseline under the same experimental setup on GSM8K and MATH. Table~\ref{tab:majority_voting} shows that CodePMP outperforms majority voting, especially on more complex tasks like MATH.

\begin{table}[!t]
    \centering
    \begin{tabular}{l|cc}
    \toprule
    \textbf{Method} & \textbf{GSM8K} & \textbf{MATH} \\
    \midrule
    CodePMP & 0.8484 & 0.41 \\
    Majority Voting & 0.8453 & 0.37 \\
    \bottomrule
    \end{tabular}
    \caption{Comparison of CodePMP and majority voting on GSM8K and MATH.}
    \label{tab:majority_voting}
\end{table}

\subsection{Sample Efficiency Improvements on Reclor and LogiQA}
We finetune the RM on preference pairs using only Reclor or LogiQA and then evaluate them on their respective test sets. As shown in Figures \ref{rm_acc_reclor_or_logiqa} and \ref{rm_bon_reclor_or_logiqa}, PMP demonstrates a clear advantage in sample efficiency, reflected in both RM accuracy and Best-of-N evaluation. The results reveal that even with substantially fewer training samples, reward models initialized with CodePMP achieve comparable or better performance than models trained from scratch with many more samples, highlighting the significant sample efficiency benefits of our approach for logical reasoning tasks.

\subsection{Performance on Coding Tasks}
\label{sec:performance_coding}

We evaluate CodePMP's effectiveness on actual code generation tasks by conducting two types of evaluations: reward model accuracy assessment and code generation evaluation. 

First, we assess the reward model's accuracy on the CodeUltraFeedback benchmark, which consists of preference pairs in the code domain. We fine-tuned Qwen2 models on the CodeUltraFeedback\_binarized dataset (8.5k preference pairs), both with and without CodePMP initialization. Table~\ref{table:code_bench_results} presents the accuracy results across model sizes.

\begin{table}[!t]
\begin{center}
\begin{small}
\begin{sc}
\begin{tabular}{cc|c}
\toprule
\multirow{2}{*}{\textbf{Model}} & \multirow{2}{*}{\textbf{PMP}} & \textbf{CodeUltraFeedback} \\
 & & \textbf{Accuracy} \\ \midrule
\multirow{2}{*}{1.5B} & \ding{55} & 0.6841 \\ \cline{2-3} \addlinespace[3pt]
 & \ding{51} & \textbf{0.758} \\ \midrule
\multirow{2}{*}{7B} & \ding{55} & 0.6912 \\ \cline{2-3} \addlinespace[3pt]
&  \ding{51} & \textbf{0.7619} \\ \bottomrule
\end{tabular}
\end{sc}
\end{small}
\end{center}
\caption{Performance on CodeUltraFeedback benchmark shows that CodePMP improves in-domain code reward modeling.}
\label{table:code_bench_results}
\end{table}

As shown in Table~\ref{table:code_bench_results}, reward models initialized with CodePMP consistently outperform those without PMP initialization on the CodeUltraFeedback benchmark. For the 1.5B model, CodePMP initialization improves accuracy from 0.6841 to 0.758, while for the 7B model, accuracy increases from 0.6912 to 0.7619. These results demonstrate that CodePMP effectively enhances reward models' ability to evaluate code quality.

Beyond reward model evaluation, we also assess whether this improved evaluation capability translates to better code generation outcomes using the HumanEval benchmark. For this evaluation, we used deepseek-coder-6.7b-instruct as the generator and Qwen2-7B as the reward model (RM). We fine-tuned the RM on the same CodeUltraFeedback\_binarized dataset, both with and without CodePMP initialization. Table~\ref{tab:human_eval} presents Pass@1 (0-shot) results under different $N$ values.

\begin{table}[!t]
    \centering
    \begin{adjustbox}{max width=\columnwidth}
    \begin{tabular}{c|c|c}
    \toprule
    \textbf{BoN} & \textbf{Qwen2-7B w/o PMP} & \textbf{Qwen2-7B w/ PMP} \\
    \midrule
    N=1   & 0.7134 & 0.7134 \\
    N=2   & 0.7317 & 0.7195 \\
    N=4   & 0.7073 & 0.7622 \\
    N=8   & 0.6890 & 0.7683 \\
    N=16  & 0.6951 & 0.7256 \\
    N=32  & 0.6585 & 0.7378 \\
    N=64  & 0.6829 & 0.7134 \\
    N=128 & 0.6707 & 0.7012 \\
    N=256 & 0.6707 & 0.7195 \\
    \bottomrule
    \end{tabular}
    \end{adjustbox}
    \caption{HumanEval results (Pass@1, 0-shot) for different numbers of sampled solutions $N$. The generator is deepseek-coder-6.7b-instruct.}
    \label{tab:human_eval}
\end{table}

The results in Table~\ref{tab:human_eval} indicate that CodePMP initialization provides a generally more stable and higher-accuracy selection mechanism compared to direct training, especially as $N$ varies. For most values of $N$, the model with CodePMP initialization achieves better Pass@1 scores, with particularly notable improvements at $N=4$ (0.7622 vs. 0.7073), $N=8$ (0.7683 vs. 0.6890), and $N=32$ (0.7378 vs. 0.6585). Without CodePMP, we observe performance degradation at higher $N$ values, while CodePMP-initialized models maintain more consistent performance. This finding is particularly significant since HumanEval evaluates actual code generation rather than just preference prediction, demonstrating that the benefits of CodePMP extend beyond improved preference modeling to better code generation outcomes.

\paragraph{Summary} 
These additional experiments demonstrate that CodePMP:
\begin{itemize}
    \item Outperforms majority voting in both simpler (GSM8K) and more challenging (MATH) settings.
    \item Demonstrates significant sample efficiency improvements on logical reasoning tasks (Reclor and LogiQA), with models initialized with CodePMP achieving better performance with fewer training samples.
    \item Provides more stable and accurate code evaluation on HumanEval, showing benefits for practical code generation tasks.
\end{itemize}

Thus, CodePMP provides a scalable and effective approach to improving large language models across different domains and tasks.

\section{Comprehensive Data Diversity Analysis}\label{appendix:data_diversity}
To validate the quality of our synthetic data, we conducted comprehensive diversity analyses using established methodologies from the research on synthetic text data generation~\citep{zhu2024synthesizetextdatamodel}. These analyses aim to demonstrate that our synthetic data maintains sufficient diversity while effectively capturing the distributions present in human-generated data.

\subsection{N-gram Feature Distribution Analysis}
\begin{figure}[ht]
    \centering
    \includegraphics[width=\columnwidth]{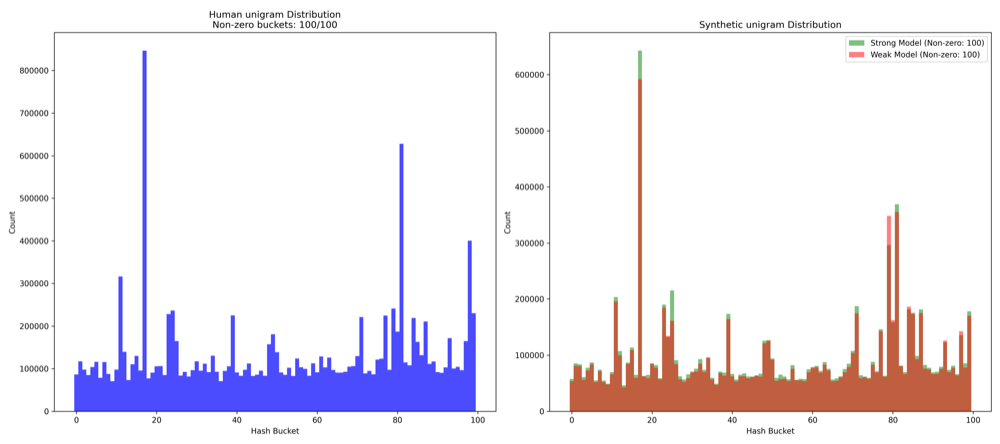}
    \caption{Unigram distribution comparison (left: distribution for human data, right: distribution for synthetic data).}
    \label{fig:unigram_distribution}
\end{figure}

\begin{figure}[ht]
    \centering
    \includegraphics[width=\columnwidth]{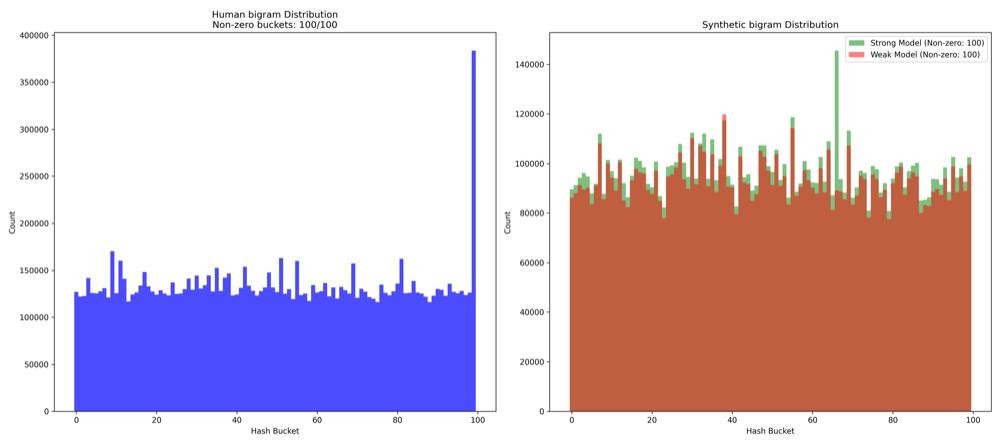}
    \caption{Bigram distribution comparison (left: distribution for human data, right: distribution for synthetic data).}
    \label{fig:bigram_distribution}
\end{figure}

We mapped text n-gram features to fixed hash buckets (100 buckets) and analyzed their distribution patterns to measure lexical diversity. Figures \ref{fig:unigram_distribution} and \ref{fig:bigram_distribution} show the comparison between human-generated data and our synthetic data.

Table \ref{tab:ngram_density} presents the density values for unigram and bigram distributions across different data sources.

\begin{table}[ht]
\centering
\begin{adjustbox}{max width=\columnwidth}
\setlength{\tabcolsep}{3pt}
\begin{tabular}{lcc}
\hline
\textbf{Data Source} & \textbf{Unigram} & \textbf{Bigram} \\
\hline
Human & 134,538.40 & 133,538.41 \\
Strong Model & 97,653.69 & 96,653.70 \\
Weak Model & 93,691.69 & 92,691.70 \\
\hline
\end{tabular}
\end{adjustbox}
\caption{N-gram density values for human and synthetic data.}
\label{tab:ngram_density}
\end{table}

The distribution graphs show that, compared to human data, our synthetic data has more uniform n-gram distributions, without the concentration peaks common in synthetic data. The density values further quantify this advantage—synthetic text's n-gram density values (Strong Model: 97,653.69, Weak Model: 93,691.69) are significantly lower than human text (134,538.40), demonstrating more balanced distribution across hash buckets.

\subsection{Embedding Space Visualization}
\begin{figure}[ht]
    \centering
    \includegraphics[width=\columnwidth]{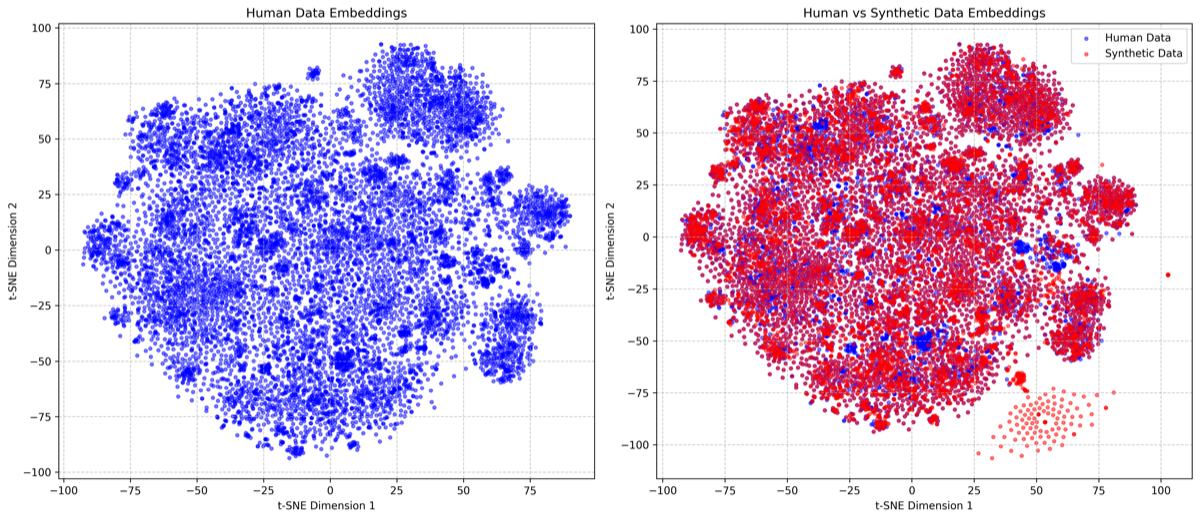}
    \caption{Embedding space visualization (left: distribution for human data, right: distribution for synthetic data).}
    \label{fig:embedding_space}
\end{figure}

To further evaluate the semantic diversity of our synthetic data, we mapped semantic features of both human and synthetic data to a 2D space, as shown in Figure \ref{fig:embedding_space}.

Both synthetic and human data show wide and dispersed distributions in the embedding space with highly overlapping distribution ranges, indicating our synthetic data captures a similarly broad semantic space as human data.

\subsection{KL Divergence Analysis}
We quantified the distribution differences between synthetic and human data using KL divergence to evaluate how closely our synthetic data approximates natural distributions. Table \ref{tab:kl_divergence} presents these results.

\begin{table}[ht]
\centering
\begin{adjustbox}{max width=\columnwidth}
\setlength{\tabcolsep}{3pt}
\begin{tabular}{lccc}
\hline
\textbf{N-gram} & \textbf{Human Internal} & \textbf{Strong Model} & \textbf{Weak Model} \\
& \textbf{(Bootstrap)} & \textbf{vs Human} & \textbf{vs Human} \\
\hline
1-gram & 0.2502 & 0.4290 & 0.4631 \\
2-gram & 0.6904 & 1.3500 & 1.4281 \\
3-gram & 1.3012 & 2.5660 & 2.6693 \\
\hline
\end{tabular}
\end{adjustbox}
\caption{KL divergence values comparing different data distributions.}
\label{tab:kl_divergence}
\end{table}

These results demonstrate that the distribution differences between our synthetic data and human data fall within acceptable ranges relative to internal human data variation.

\subsection{Comprehensive Validation of Synthetic Preference Data}
Our synthetic data generation approach relies on two key assumptions: (1) larger models from the same family produce higher-quality code than smaller ones, and (2) this quality difference creates consistent preference signals suitable for training. We conducted both theoretical and empirical validation to confirm these assumptions.

\subsubsection{Validation of Strong-Weak Model Ability Differences}
To validate our first assumption, we analyzed ability differences between strong and weak models across various code-related dimensions.

\begin{figure}[ht]
    \centering
    \begin{subfigure}[b]{0.48\columnwidth}
        \centering
        \includegraphics[width=\columnwidth]{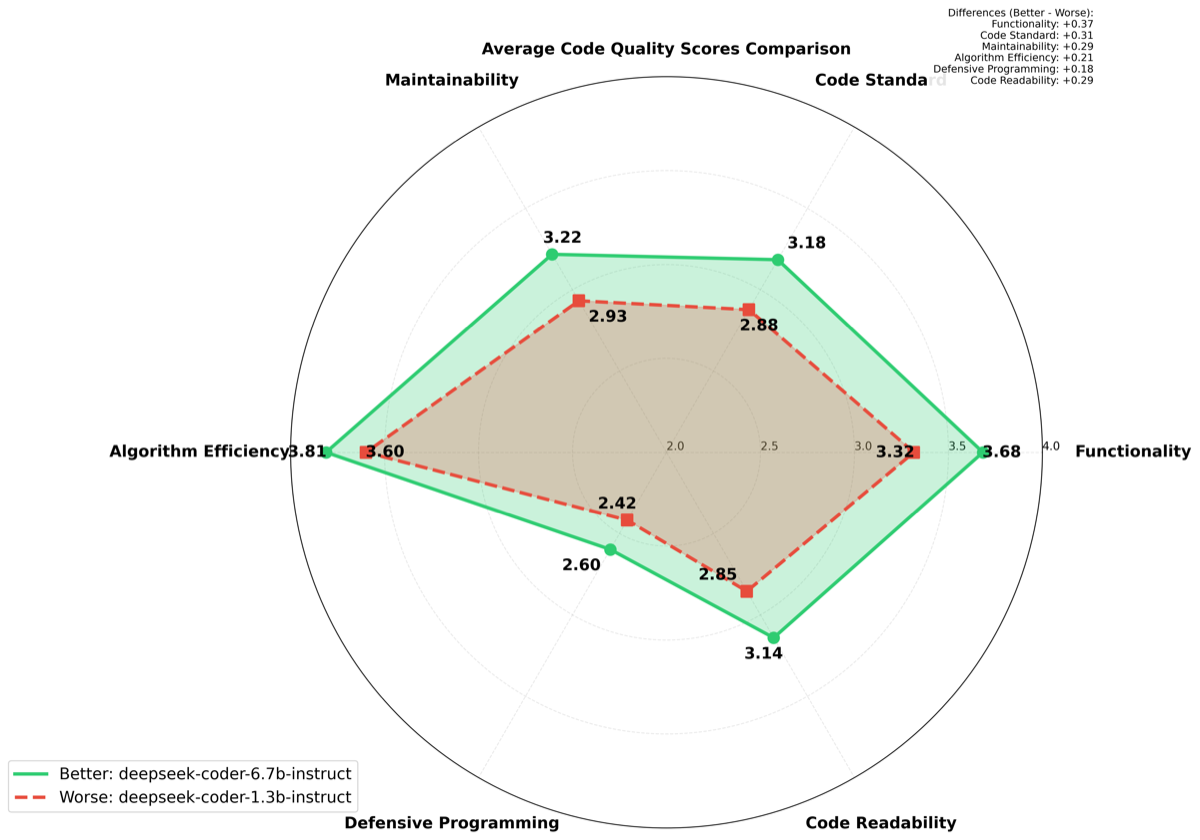}
        \caption{Same model family}
        \label{fig:radar_same}
    \end{subfigure}
    \hfill
    \begin{subfigure}[b]{0.48\columnwidth}
        \centering
        \includegraphics[width=\columnwidth]{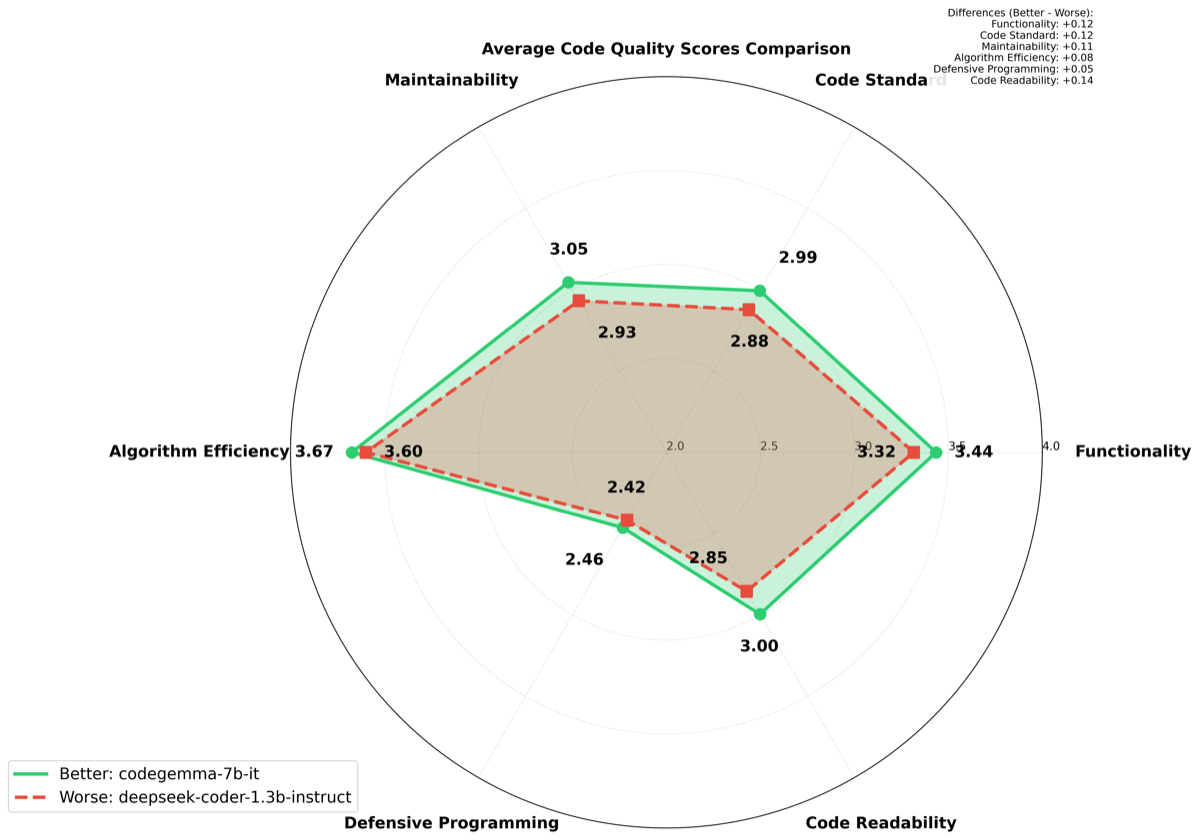}
        \caption{Different model families}
        \label{fig:radar_diff}
    \end{subfigure}
    \caption{Radar charts showing ability differences between strong and weak LLMs across various dimensions.}
    \label{fig:model_family_comparison}
\end{figure}

Figure \ref{fig:radar_same} provides strong evidence supporting our assumption: when using models from the same architectural family with different parameter counts, the stronger model consistently outperforms the weaker model across all ability dimensions. This uniform superiority ensures that synthetic preference pairs have clear and consistent quality differences, creating reliable signals for training preference models.

For comparison, Figure \ref{fig:radar_diff} shows what happens when models from different families are paired. Here, we observe irregular and inconsistent differences, with some dimensions showing negligible gaps or even inversions. Such inconsistencies could potentially introduce noise into the preference signals, undermining training data quality.

\subsubsection{External Evaluation of Preference Consistency}
Having confirmed the underlying ability differences, we next validated whether these differences translate to consistent preference judgments. We conducted preference annotation experiments using GPT-4o to evaluate the consistency of preferences in our dataset. 

The results show that our synthetic CodePMP data achieved a preference consistency rate of 75.12\%. This is notably higher than the more costly CodeUltraFeedback preference dataset (71.56\%), demonstrating that the preference distinction in our synthetic data (based on our assumption that "larger models generate better code than smaller models") is sufficiently clear and consistent.

This external validation reinforces our first finding - not only do larger models from the same family consistently outperform smaller ones across all dimensions, but this performance gap is readily detectable by strong evaluator models, resulting in consistent preference judgments.

As CodePMP is fundamentally a pretraining process, we deliberately simplified our assumptions to enable scalable preference data creation with minimal additional validation. Our multi-faceted validation approach confirms that this simple yet effective methodology produces high-quality, consistent preference data suitable for large-scale pretraining.

\subsection{Source Data Quality and Experimental Validation}
Our method achieves excellent diversity due to the high quality of our source data:
\begin{itemize}
    \item We collected over 130 million code snippets from GitHub, covering all common programming languages and task types on open-source platforms, ensuring breadth and depth in our source data.
\end{itemize}

Furthermore, our experimental results validate the effectiveness of our synthetic data diversity:
\begin{itemize}
    \item As shown in Figure \ref{fig:res_cmp_pair_construction}, our synthesis strategy outperforms preference pairs constructed directly from source code, indirectly proving that our synthesis process enhances data diversity and quality.
\end{itemize}

This comprehensive diversity analysis confirms that our synthetic data generation approach produces high-quality, diverse data that effectively captures the distribution characteristics of human-generated code. The balanced distribution patterns and semantic coverage demonstrate that our synthetic data is well-suited for training robust reward models.

\section{Detailed Implementation of CodePMP}\label{appendix:codepmp_algorithm}
In this section, we provide a detailed overview of the Code Preference Model Pretraining (CodePMP) implementation. The following description illustrates the systematic process of generating and utilizing code-preference pairs for pretraining preference models, which can then be fine-tuned for downstream reasoning tasks.

The algorithm begins with a source code repository, a strong CodeLLM (in our implementation, deepseek-coder-6.7b-instruct), and a weaker CodeLLM (deepseek-coder-1.3b-instruct). First, descriptions are generated for each code snippet in the repository using the strong model. For each description, the strong model generates a high-quality chosen response, while the weaker model generates a less optimal rejected response. These pairs are used to calculate both language modeling loss (on the responses) and reward modeling loss (comparing chosen vs. rejected responses). The final training objective combines these two loss components.

This scalable approach allows for creating millions of preference pairs without expensive human annotation, providing an effective initialization for reward models that will later be fine-tuned on specific reasoning tasks.

\begin{algorithm*}[tb]
   \caption{Code Preference Model Pretraining}
   \label{alg:code_pmp}
\begin{algorithmic}
    \REQUIRE Source code repository $S$, \\
    Strong CodeLLM $M_{\text{strong}}$, \\
    Weak CodeLLM $M_{\text{weak}}$
    \ENSURE Pretrained Model
    \STATE \textbf{Input:} Source code $S$
    \STATE Summarize description $D$ using $M_{\text{strong}}$ on $S$
    \FOR{each $D_i \in D$}
        \STATE Generate \textit{Chosen Response}  using $M_{\text{strong}}$
        \STATE Generate \textit{Rejected Response}  using $M_{\text{weak}}$
    \ENDFOR
    \STATE Calculate LM Loss $\mathcal{L}_{\text{LM}}$ on \textit{Response}
    \STATE Calculate RM Loss $\mathcal{L}_{\text{RM}}$ using \textit{Chosen Response} and \textit{Rejected Response}
    \STATE Train PMP Model using $\mathcal{L}_{\text{PMP}} = \mathcal{L}_{\text{RM}} + \mathcal{L}_{\text{LM}}$
\end{algorithmic}
\end{algorithm*}

\section{Logical Reasoning Evaluation Examples} \label{logic_reason_eval_exap}
We randomly select and present examples from the Reclor test set, which consists of multiple-choice questions based on a given passage. While it is possible to have the model generate additional candidate answers to create a Best-of-N test, it becomes difficult to ensure that the original correct answer remains among the options after introducing new candidates, and to identify the new correct answer. We attempt to use GPT-4o to annotate the correct answers for 32 responses, but the consistency with manual inspection is low, as is the consistency of GPT-4o's own multiple annotations. It can be inferred that the consistency rate would worsen if expanded to 256 responses. Therefore, after careful consideration, we decide to use RM to score only the original four manually annotated answer options, match the top-ranked option with the manually annotated correct answer, and calculate accuracy. In principle, this method is equivalent to the Best-of-4 test.

\newpage
\begin{table*}[t]
\caption{Examples from the Reclor test set, illustrating multiple-choice questions format with passages and questions.}
\setlength{\tabcolsep}{4pt}
\small
\begin{tabular}{p{0.05\textwidth} p{0.35\textwidth} p{0.45\textwidth} c}
\hline
\textbf{ID} & \textbf{Passage} & \textbf{Question} & \textbf{Ans.} \\
\hline
12824 & Mayor: When we reorganized the police department, critics claimed it would make police less responsive and lead to more crime. Statistics show an overall decrease in thefts after reorganization. & Which statement most challenges the mayor's argument?
(1) Similar reorganizations in other cities led to increased thefts.
(2) Unresponsive police reduce theft reporting rates.
(3) Critics agree police statistics are reliable.
(4) The reorganization saved less money than planned. & 2 \\
\hline
218 & Jupiter is the largest planet with mass 2.5 times that of all other planets combined. Most of Jupiter's 70+ moons are water ice. & What best supports that Jupiter's atmosphere should contain water?
(1) Satellites may eventually fall onto planets.
(2) Interstellar water exists as gas.
(3) Uranus, also a gas giant, contains water ice.
(4) Satellites and planets form from the same materials. & 3 \\
\hline
10376 & Lake Dali fish must migrate to river headwaters to breed, though no rivers connect to the sea. Scientists believe these fish originally came from the ocean. & What best explains scientists' belief?
(1) Similar fish elsewhere are larger.
(2) The fish quickly die in sea/fresh water.
(3) Lake Dali was once connected to an ocean-bound river.
(4) Fish from Lake Dali survived in far-away lakes. & 2 \\
\hline
13334 & If nuclear waste posed no threat, it could be placed in populated areas. But it is only dumped in sparsely populated regions, suggesting safety concerns. & What would most weaken this argument?
(1) Uncertain safety justifies minimal risk placement.
(2) Chemical waste is also dumped away from population.
(3) Accidents affect fewer people in sparsely populated areas.
(4) Remote locations reduce bureaucratic complications. & 3 \\
\hline
\end{tabular}
\end{table*}

\end{document}

%% file: math_commands.tex
\usepackage{amsmath,amsfonts,bm}

\def\eqref#1{equation~\ref{#1}}

\def\1{\bm{1}}

\DeclareMathAlphabet{\mathsfit}{\encodingdefault}{\sfdefault}{m}{sl}
\SetMathAlphabet{\mathsfit}{bold}{\encodingdefault}{\sfdefault}{bx}{n}

\newcommand{\E}{\mathbb{E}}

\newcommand{\R}{\mathbb{R}}

\makeatletter
\def\UrlAlphabet{%
\do\a\do\b\do\c\do\d\do\e\do\f\do\g\do\h\do\i%
\do\j\do\k\do\l\do\m\do\n\do\o\do\p\do\q\do\r%
\do\s\do\t\do\u\do\v\do\w\do\x\do\y\do\z%
\do\A\do\B\do\C\do\D\do\E\do\F\do\G\do\H%
\do\I\do\J\do\K\do\L\do\M\do\N\do\O\do\P%
\do\Q\do\R\do\S\do\T\do\U\do\V\do\W\do\X%
\do\Y\do\Z}
\def\UrlDigits{\do\1\do\2\do\3\do\4\do\5\do\6\do\7\do\8\do\9\do\0}
\g@addto@macro{\UrlBreaks}{\UrlOrds}
\g@addto@macro{\UrlBreaks}{\UrlAlphabet}
\g@addto@macro{\UrlBreaks}{\UrlDigits}
\makeatother 